\documentclass[sigconf]{acmart}

\usepackage{booktabs} 

\setcopyright{rightsretained}

\acmConference[ICVGIP'18]{11th Indian Conference on Computer Vision, Graphics and Image Processing}{Dec. 2018}{Hyderabad, India}
\acmYear{2018}
\copyrightyear{2018}

\acmArticle{4}
\acmPrice{15.00}

\copyrightyear{2018}
\acmYear{2018}
\setcopyright{acmlicensed}
\acmConference[ICVGIP 2018]{11th Indian Conference on Computer Vision, Graphics and Image Processing}{December 18--22, 2018}{Hyderabad, India}
\acmBooktitle{11th Indian Conference on Computer Vision, Graphics and Image Processing (ICVGIP 2018), December 18--22, 2018, Hyderabad, India}
\acmPrice{15.00}
\acmDOI{10.1145/3293353.3293367}
\acmISBN{978-1-4503-6615-1/18/12}

\editor{Anoop M. Namboodiri}
\editor{Vineeth Balasubramanian}
\editor{Amit Roy-Chowdhury}
\editor{Guido Gerig}

\begin{document}
\title{Weakly Supervised Object Localization on grocery shelves using simple FCN and Synthetic Dataset}
\renewcommand{\shorttitle}{WSOL using simple FCN and Synthetic Dataset}

\author{Srikrishna Varadarajan}
\authornote{Both authors contributed equally}
\affiliation{%
  \institution{Paralleldots, Inc.}
}
\email{srikrishna@paralleldots.com}

\author{Muktabh Mayank Srivastava\footnotemark[1]}
\affiliation{
  \institution{Paralleldots, Inc.}
}
\email{muktabh@paralleldots.com}



\begin{abstract}
We propose a weakly supervised method using two algorithms to predict object bounding boxes given only an image classification dataset. First algorithm is a simple Fully Convolutional Network (FCN) trained to classify object instances. We use the property of FCN to return a mask for images larger than training images to get a primary output segmentation mask during test time by passing an image pyramid to it. We enhance the FCN output mask into final output bounding boxes by a Convolutional Encoder-Decoder (ConvAE) viz. the second algorithm. ConvAE is trained to localize objects on an artificially generated dataset of output segmentation masks. We demonstrate the effectiveness of this method in localizing objects in grocery shelves where annotating data for object detection is hard due to variety of objects. This method can be extended to any problem domain where collecting images of objects is easy and annotating their coordinates is hard.
\end{abstract}

%
%



\maketitle

\section{Introduction}

Visual retail audit or shelf monitoring is an upcoming area where computer vision algorithms can be used to create automated system for recognition, localization, tracking and further analysis of products on retail shelves. Our work focuses on the task of product localization on shelves, i.e object detection. We approach the task as weakly supervised object detection and our method ends up solving the task of product image recognition (classification) too as a subtask.

Manual annotation of object bounding boxes across images having a large number of instances per image and/or a large number of object categories is both a time consuming and an extremely costly process. One can observe that these numbers are low for most publicly available object detection datasets (average instances per image is less than 3 for PASCAL VOC and ImageNet and 7.7 for MS-COCO, number of classes is 20 for PASCAL VOC, 200 for ImageNet, 80 for MS-COCO). Annotating retail product bounding boxes in images is infeasible due to number of retail products available, frequent changes in packaging, introduction of new products (large number of classes and dynamic appearance) and high density with which objects are placed in shelves (large number of instances per image). This difficulty in acquiring annotated dataset limits the use of fully supervised object detectors in such datasets.

To perform object detection tasks in such datasets, we need a method which uses just object instances and not only learns to recognize the instances, but also localize them. Our work proposes a method for the same. Methods like ours are classified into weakly supervised object detection methods as only (few) instances are available as opposed to fully supervised object detection methods which work on multiple images with bounding box tagged objects.

We aim to do object localization using the same network which is trained for the task of classification on object instances. The classifier we use here is a Fully Convolutional Network (FCN) \cite{NIN}. All the learnable layers in FCN are convolutional layers. There are no fully connected (dense) layers in the network. This allows us to pass an input of any higher resolution to the network and get a corresponding output mask. We also make use of another optional network to enhance our outputs from the FCN for our object detection task. For this network (Refine-net), we use a convolutional network based on the encoder-decoder framework which we call as ConvAE. ConvAE is trained on an artificial dataset created using the object instances. The output of the ConvAE is converted to obtain bounding boxes for the object detection task. Please note that we use object detection and object localization invariably throughout this paper.

\begin{figure*}
\centering
\includegraphics[width=0.95\textwidth]{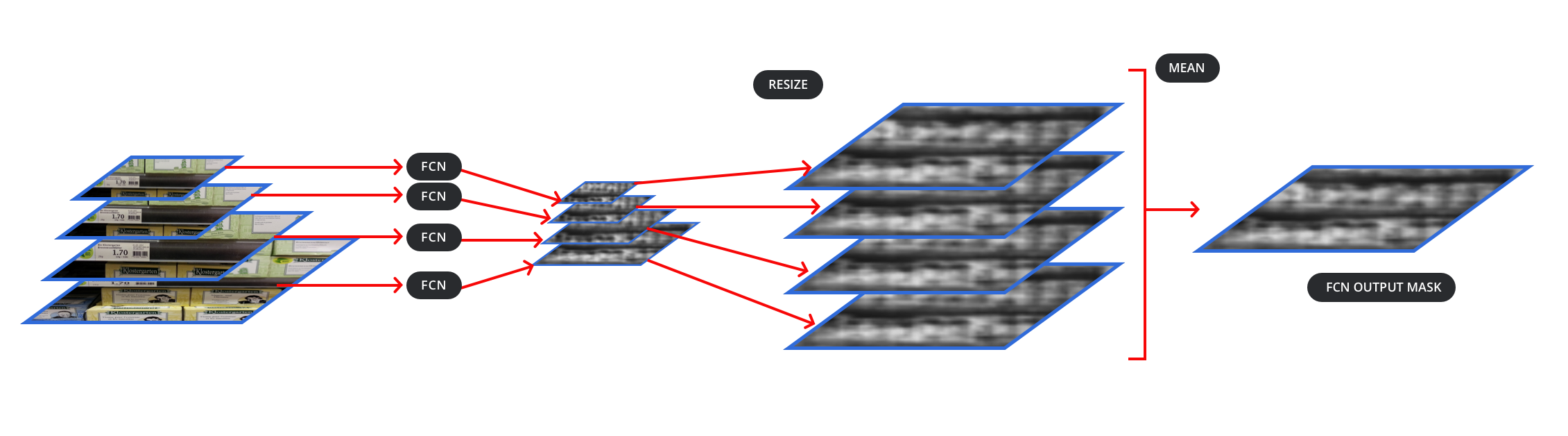}
\caption{Object detection pipeline on shelf images}
\label{fig:testing}
\end{figure*}

\section{Weak Supervision and Synthetic Datasets}

Weakly supervised learning is a method of learning in which the supervision can be indirect, inexact and incomplete. In this setting, one doesn't have access to supervised data but some other form of labeled data which is useful. There have been previous works \cite{progressive} \cite{selftaught} \cite{isOLfree} \cite{attentionWSL} \cite{hidenseek} which does the weakly supervised localization (WSL) task on the standard datasets like PASCAL VOC and COCO. The WSL methods are usually Multiple Instance Learning (MIL) based or end-to-end CNN based networks. Our method falls into the latter category. One could use the visual explanation method described in \cite{gradcam} to do the same WSOL task, but converting the attention maps to bounding boxes fails in the case of multiple (small) objects which are kept together (as seen in retail shelves).

The FCN component of our method shares some conceptual similarity with \cite{isOLfree}, where a classifier trained on instances is used for object detection too. Our model is simpler and doesn't use multi-scale sliding windows. Our model also gives perfect bounding boxes unlike the point (center) output from latter. There is also difference amongst the datasets both the algorithms are trained/tested on, while we train it on single label object images for retail objects, the latter algorithm has been trained on multilabel datasets like tags of PASCAL VOC.

Our optional refinement module is trained on a Synthetic Dataset. The creation of Synthetic Dataset has been explored in the past for various object detection tasks \cite{syntheticindoor} \cite{logo} \cite{virtual}. Some use sophisticated methods to place the objects in its environments to expand an existing dataset or create a new one, while we just simply place objects on empty shelves with few logical parameters for randomness. The novelty of our approach is that the optional refinement module is \textbf{not trained on directly synthesized images} of Synthetic Dataset, but rather their \textbf{output after passing through FCN}. Moreover, we use the Synthetic Dataset only to improve the extent of our localization and to reduce false positives.

\section{Method}
\label{method}

\subsection{FCN}

\subsubsection{FCN: Training}

\begin{figure}
\centering
\includegraphics[width=0.5\textwidth]{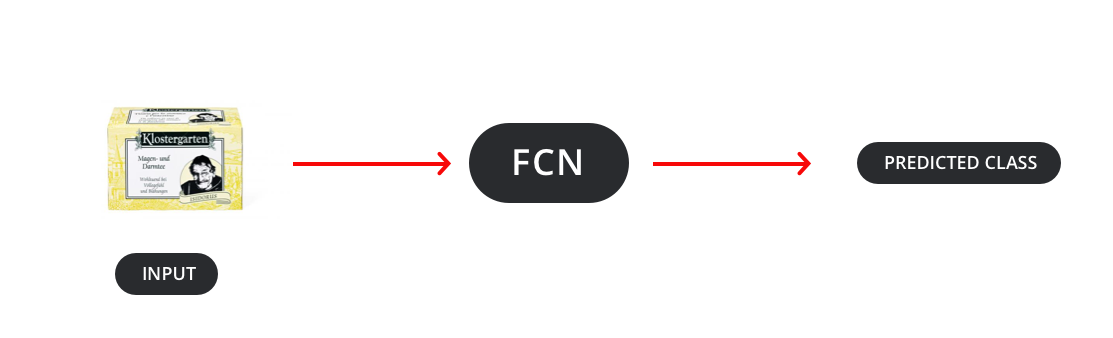}
\caption{Classification Pipeline on object instances}
\label{fig:cls}
\end{figure}

During training task, we perform classification on instances of the objects. The network architecture of the FCN is as follows. We take the pretrained layers of VGG11\cite{vgg} from torchvision\footnote{\url{https://github.com/pytorch/vision/blob/master/torchvision/models/vgg.py}} and remove all the FC Layers. We then add a single convolutional layer at the end with appropriate kernel size so as to get a Nx1x1 output, where N is number of classes. This is shown in Figure \ref{fig:cls}.

\subsubsection{FCN: Testing}

One of the core advantages of our method is that FCNs are independent of input size. We take the valid assumption that the object instances (training images) are smaller in size than testing images (images containing multiple objects). Hence, the FCN which is designed to do a classification task (1x1 output) will give a corresponding output mask when a higher resolution input is passed. 

In the testing task, we pass an image pyramid (consisting of various downscales of the image) to the same network. We use downscale factor of 1.5 to generate the pyramid. We resize the corresponding outputs from each level of pyramid to the original image size. We then take the mean of the pyramid outputs as our final output mask as shown in the Figure \ref{fig:testing}.

Since the network is trained on individual instances of the object, it's possible that the dataset doesn't capture the various sizes of the object. This can be a huge drawback during test time. But this is overcome by the image pyramid method employed during test time. The idea is that network would be able to capture the object at least at one downsampled level.

\subsection{Refine-Net}

Refine-Net module is used for refining the outputs of FCN during object detection task. 
FCN output maps capture the objects, but they focus on more discriminative areas as they were trained for classification. This can localize the discriminative features of objects, but cannot always give bounding boxes surrounding them. As a result it might give low IoU scores. FCNs also may have false positives on background (due to lack of background training data). Refine-Net is hence used to address both these limitations. It is convolutional network which is trained to improve the output maps generated by the FCN. Please note this is an optional module and is not needed in case FCN works well on its own. One can see the performance differences later in the results section.

\begin{figure}
\centering
\includegraphics[width=0.5\textwidth]{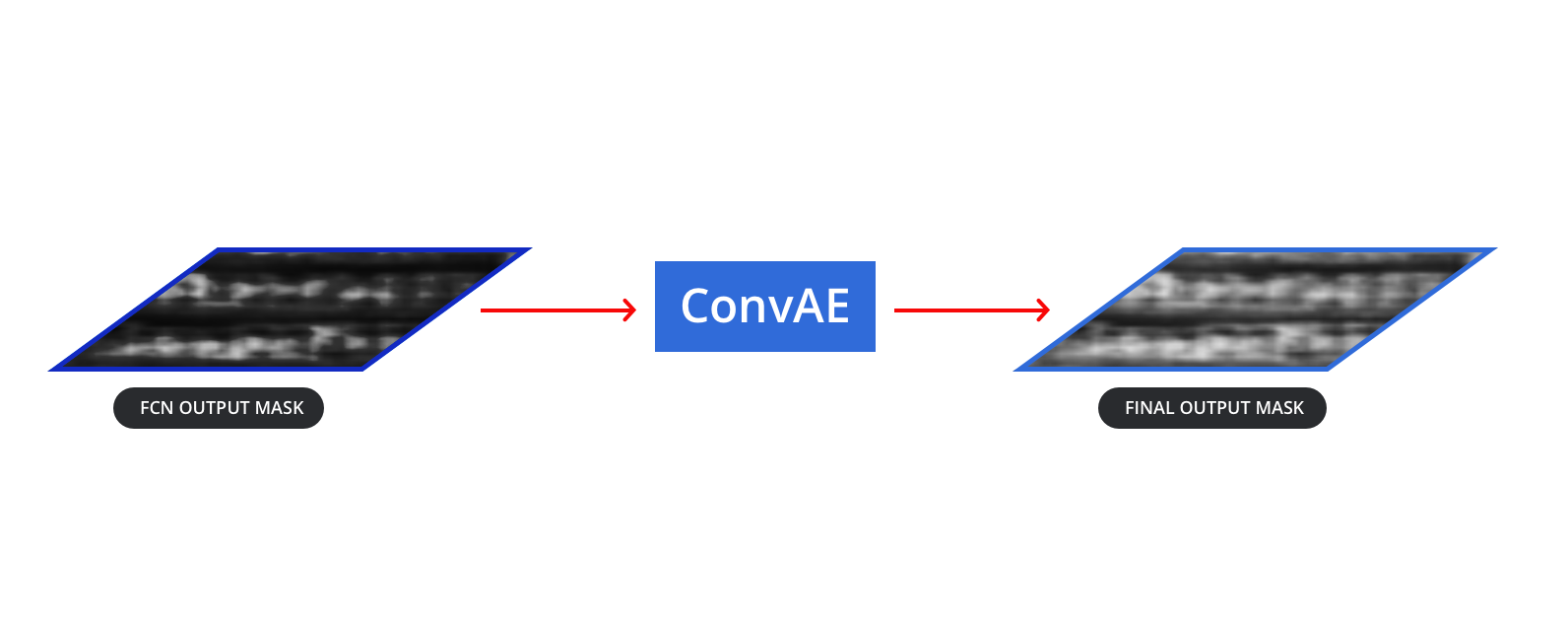}
\caption{Refinement Module}
\label{fig:convae}
\end{figure}

\subsubsection{Synthetic Dataset}

For this task, we create a synthetic dataset using the object instances from the trainset. We randomly take few images (20-50) from Google Images using the keyword \textit{empty shelves}. On these empty shelf images, we randomly place objects in a planogram format with varying number of columns (within each image) and varying number of rows (across images). We also scale the object sizes randomly between logical values. Eg: 0.5-1.1. We also tried this with a black background instead of empty shelves. We see a small increase in the mAP when we use the empty shelves.

\subsubsection{Refine-Net: Architecture}

\begin{figure}
\centering
\includegraphics[height=0.45\textheight,width=0.5\textwidth]{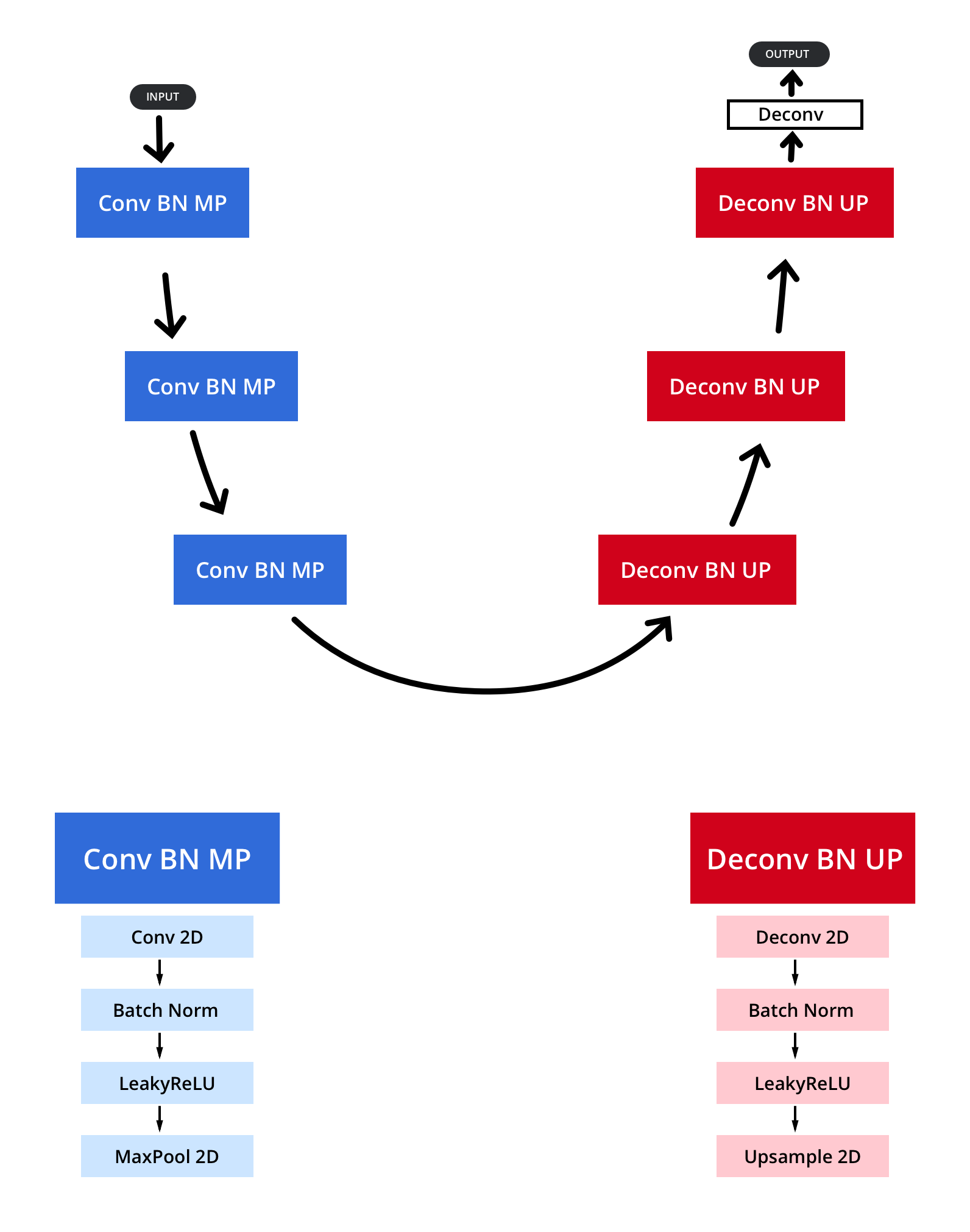}
\caption{Refine-net architecture}
\label{fig:convae_arch}
\end{figure}

We name the architecture used as Refine-Net as ConvAE. The ConvAE network follows the standard Encoder-Decoder framework with few layers of Convolutional and Downsampling layers followed by equivalent number of Deconvolutional and Upsampling layers. All the Conv2D and Deconv2D layers have 3x3 kernel size except the final Deconv which is 1x1. All MaxPool2D layers have kernel size 2x2 while all the Upsample layers are bilinear and have factor as 2. The architecture is illustrated in the Figure \ref{fig:convae_arch}.

\subsubsection{Refine-Net: Method}

During training, the images from synthetic dataset are passed through the FCN and the segmentation output is obtained. The FCN output mask is the input to the ConvAE (Refine-Net). We train the ConvAE with CrossEntropyLoss and ground-truth which is known to us from the Synthetic Dataset. 

\begin{figure*}
\includegraphics[width=0.49\textwidth, height=0.2\textheight]{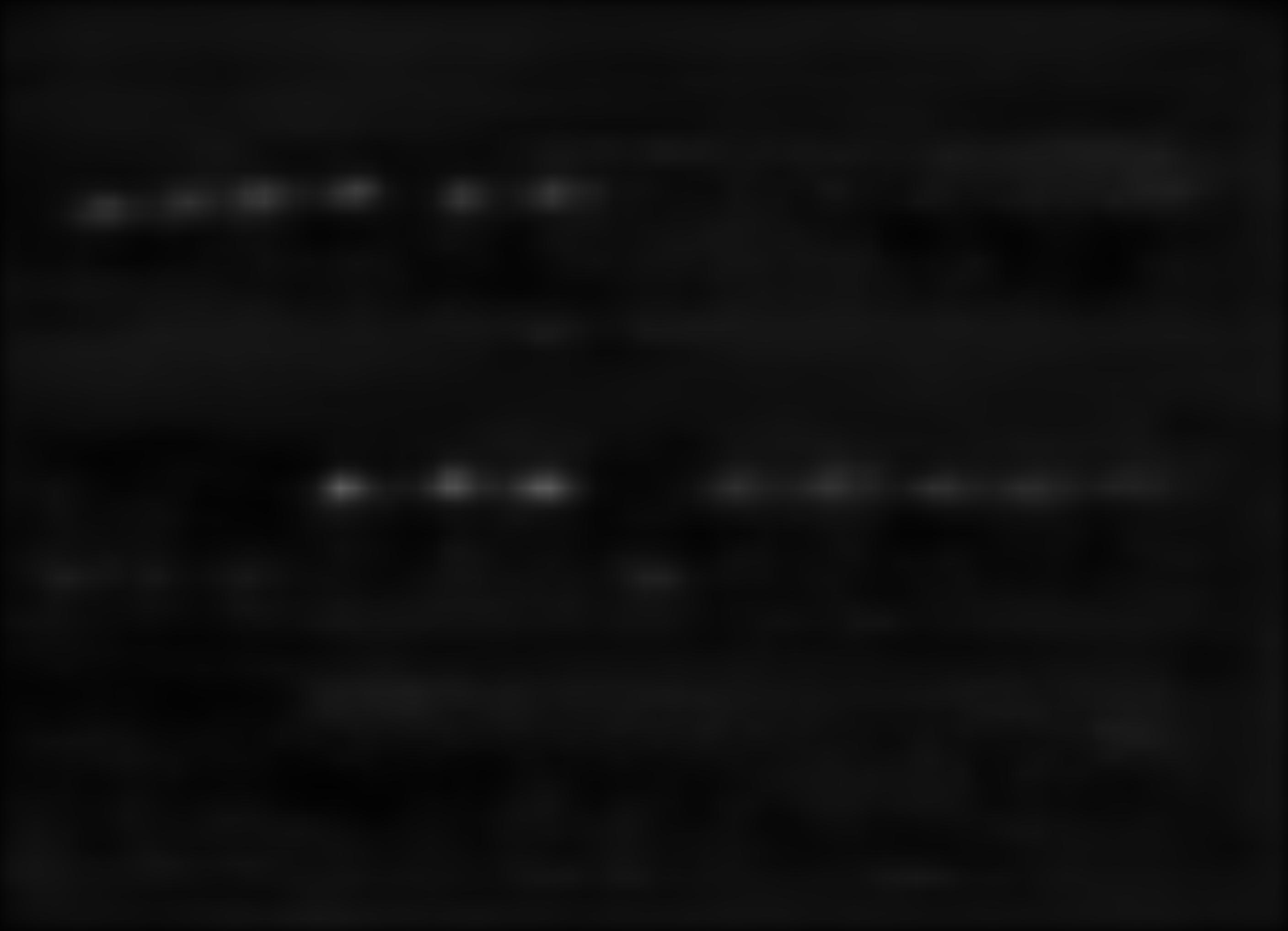}
\includegraphics[width=0.49\textwidth, height=0.2\textheight]{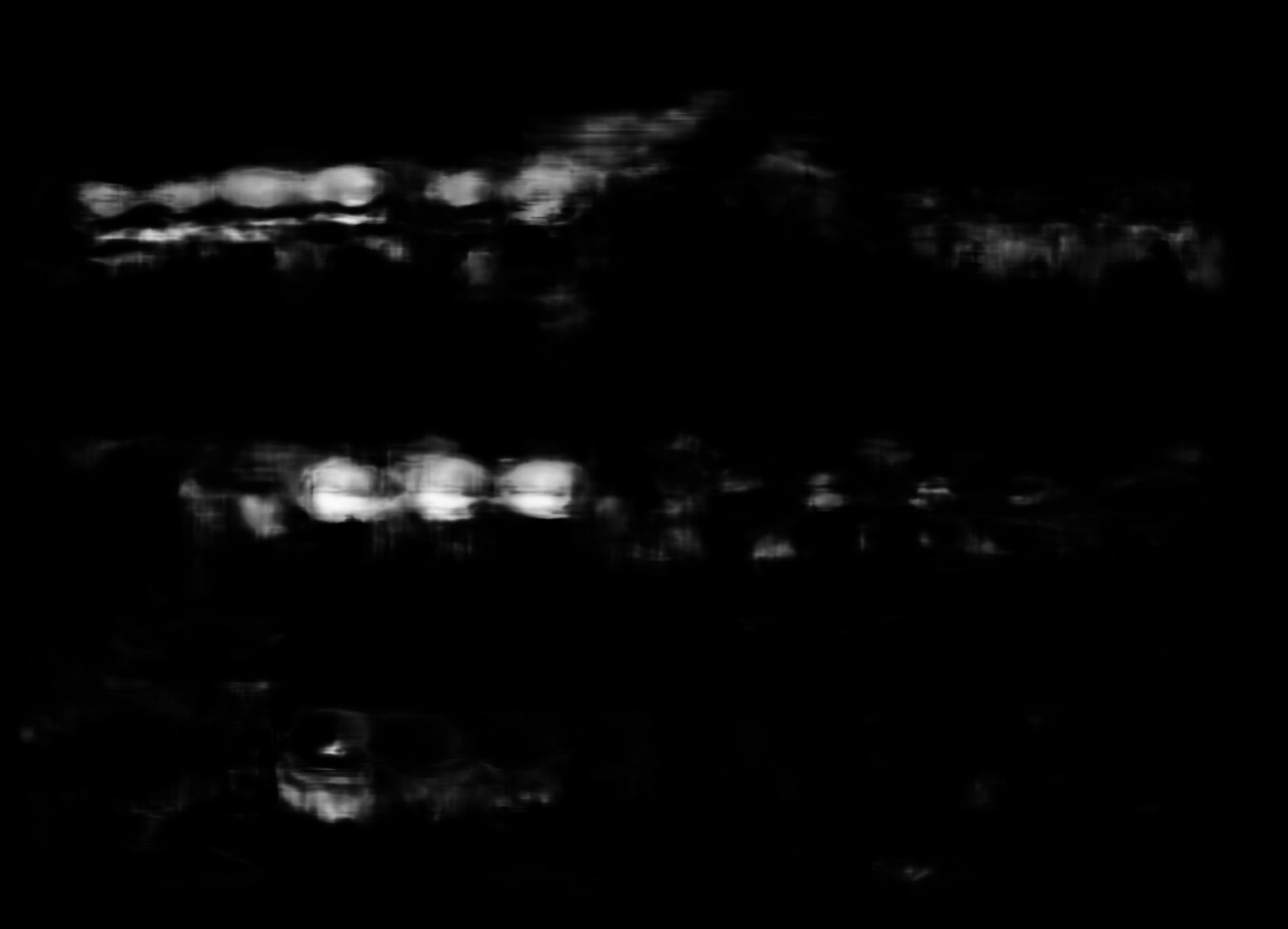}
\includegraphics[width=0.49\textwidth, height=0.2\textheight]{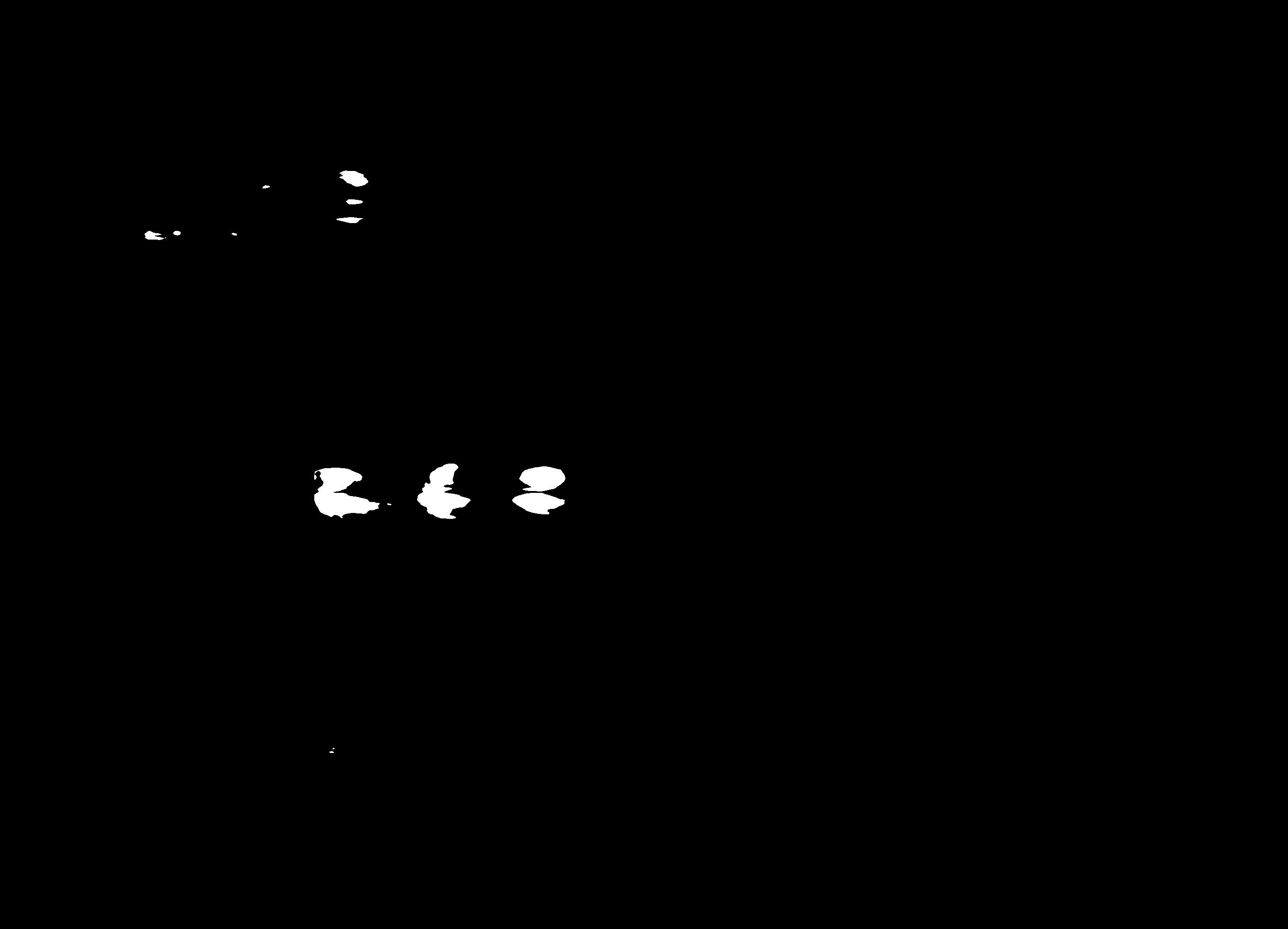}
\includegraphics[width=0.49\textwidth, height=0.2\textheight]{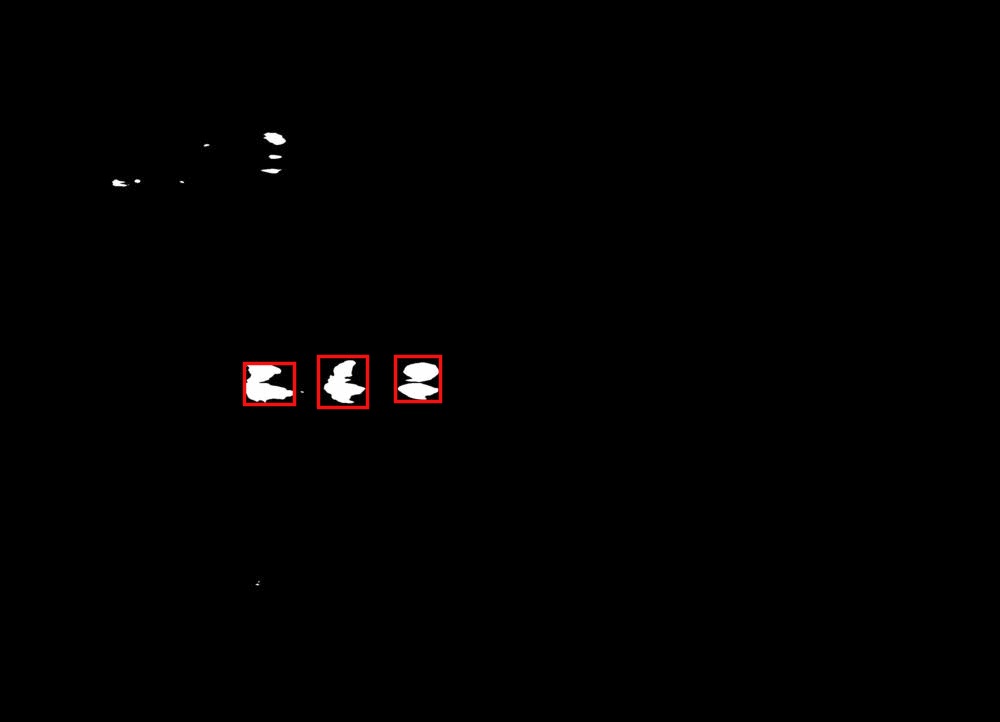}
\caption{From top: Output from FCN, Output from ConvAE,
		Output after thresholding, Output after creating bounding boxes}
\label{fig:all_ops}
\end{figure*}

During testing, the output of the FCN is passed through ConvAE to be enhanced and made better for localization. The output segmentation mask of ConvAE is now processed for extracting bounding boxes. We first binarize the mask by applying a threshold. Then, extract the bounding box coordinates from the connected components of the binarized mask. This process is shown in Figure \ref{fig:all_ops}.

\subsection{Observation}
\label{observation}

In the datasets of product recognition, it often happens that there are very similar products which belong to different classes. This tends to make the classifier's (FCN) output probability of the correct class a little weaker. Here, the confidence score of the correct class isn't as high as we'd like it to be. In such cases, one can find reasonable confidence score in \textbf{other classes} even though the FCN predicted the correct class\footnote{gave maximum probability for the correct class}. Hence, this induces a lot of false-positives depending on the \textbf{threshold} of converting the segmentation into a binary mask. 

The ConvAE works great in eliminating these false-positives. It is able to learn the prominent class and remove the false-positives from other classes easily. Hence we don't have to worry about the threshold\footnote{Note that in this case, the classifier predicts the correct class, just the confidence score probabilities at the spatial location is lesser}. We note that it's not possible to train the ConvAE if our classifier isn't strong. The false positives makes it very difficult for the network to learn the refinement task.

\section{Experiments}

We were interested in applying this method to a grocery shelf dataset. We also wanted to test the performance of the method in detecting different products from the same brand which look very similar. We were able to find one such public dataset \cite{cigarette} which is most similar to our objective, we're calling it the Cigarette Dataset. We also wanted to show the performance of the method in a dataset which is not small and controlled but bigger and more generalized. Hence, we chose the Grocery Dataset \cite{grocerydset}. The dataset details, experiments and results on both the datasets are explained in the following sections

\section{Cigarette Dataset}

\subsection{Dataset description}

It's important to explain the dataset to understand our experiments and results. This dataset consists of various products from multiple brands of cigarettes. The dataset considers 10 brands as positive classes and the rest as background. The dataset consist of 3 parts
  \begin{enumerate}
  \item Product Images
  \item Brand Images
  \item Shelf Images
  \end{enumerate}

\textbf{Product Images} are the instances of products of each brand. It contains 5 unique product images taken by 4 different cameras in different lighting, angles and noisy versions making the total number to be around 350 for each product of a brand. \textbf{Brand Images} are taken directly from the Product Images by cropping out only the brand logos from the cigarette boxes. There are 354 \textbf{Shelf Images} consisting of grocery shelf images. It contains various cigarette product boxes from different brands. There are 13,000 products in Shelf Images overall and all of them are annotated. Around 3000 belong to the 10 classes, the rest 10,000 products are considered as negative or background class.  

\begin{figure}
\centering
\includegraphics[width=0.5\textwidth, height=0.2\textheight]{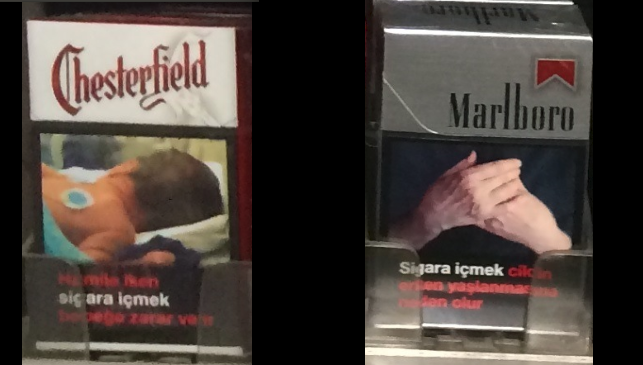}
\caption{Negative examples: Belongs to background class}
\label{fig:neg}
\end{figure}

\begin{figure}
\centering
\includegraphics[width=0.3\textwidth, height=0.2\textheight]{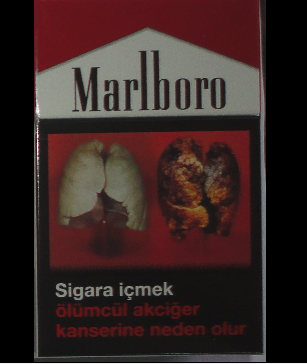}
\caption{Positive example: Belongs to Marlbaro class}
\label{fig:pos}
\end{figure}

Negative object instances contains products which don't belong to the 10 brands as well as products which belong to the 10 brands but are not the products in our 10 classes. This is explained in Figure \ref{fig:neg}. Chesterfield (brand) doesn't belong to any of the 10 brands and is a negative class while Marlboro (brand) is one of the 10 brands but the given instance is a different product than those in our trainset. Hence, this is also a negative class. A positive example of Marlboro class can be seen in Figure \ref{fig:pos}.

\begin{figure*}
\includegraphics[width=0.49\textwidth, height=0.2\textheight]{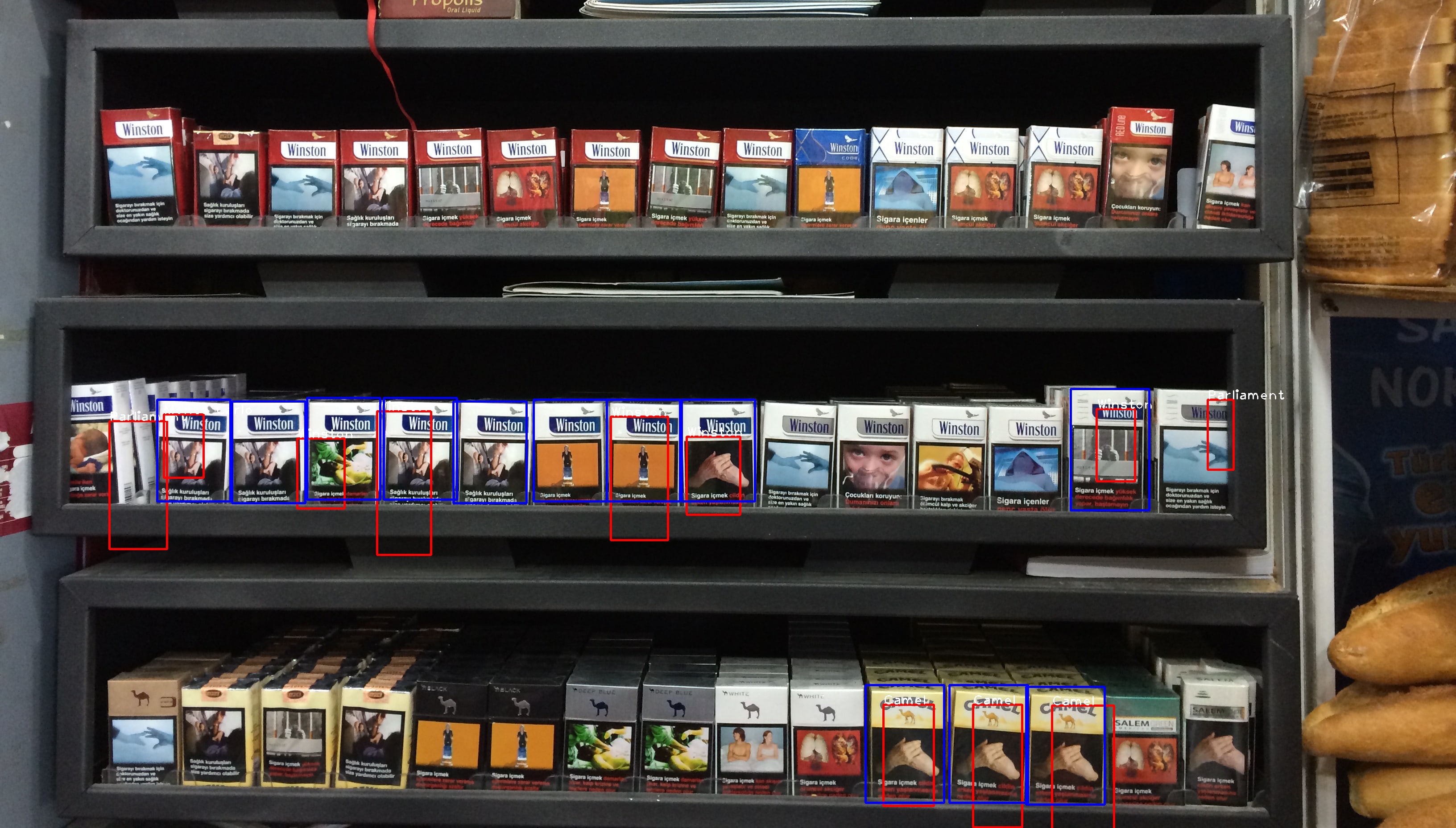}
\includegraphics[width=0.49\textwidth, height=0.2\textheight]{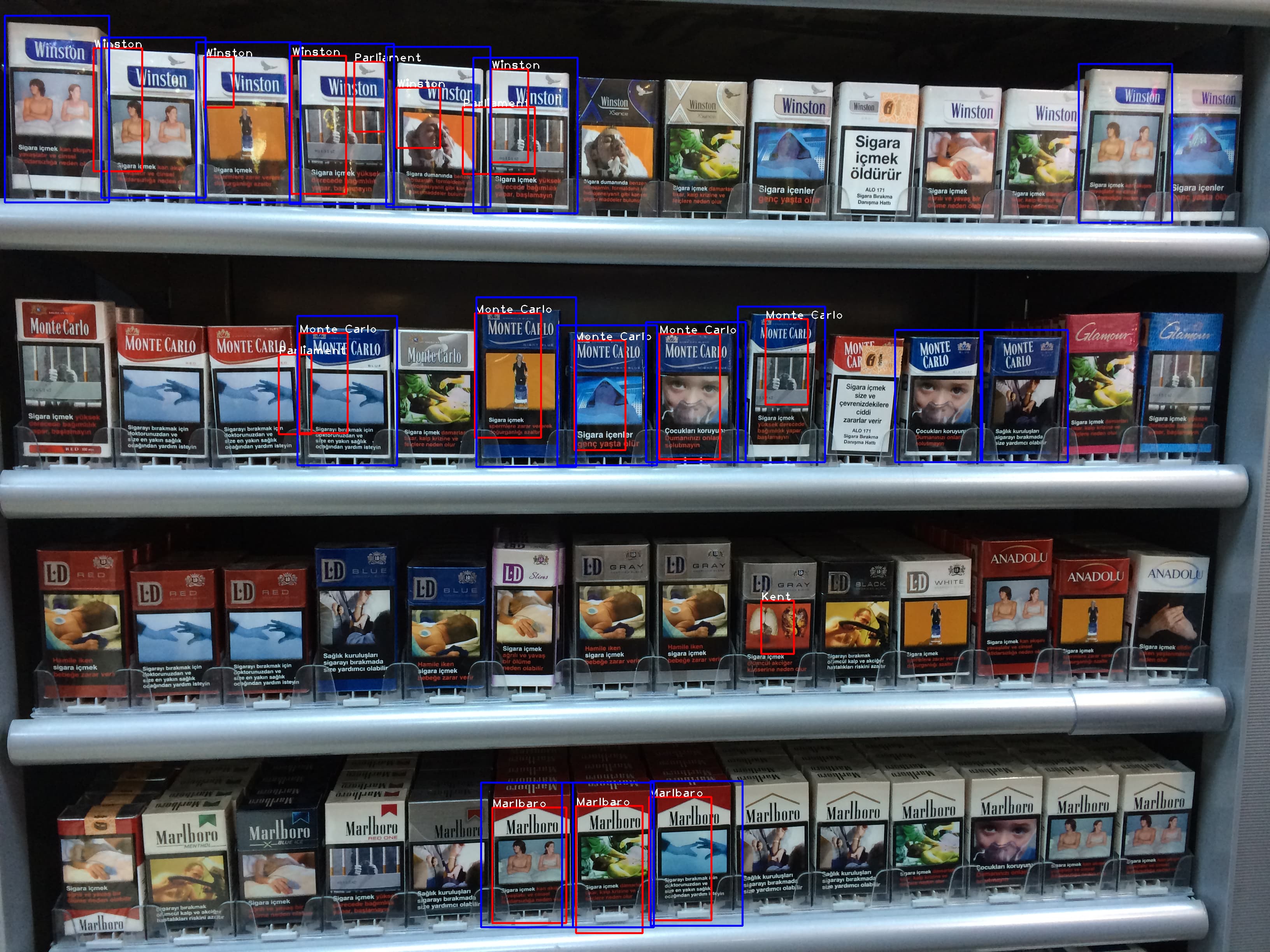}
\caption{Sample Results of Pyramid FCN Model with ConvAE. Blue shows ground-truth boxes while Red shows model's prediction}
\label{fig:cigresult}
\end{figure*}


\subsection{Train set}
The training task is the classification task. For our experiment, we chose the Brand Images as the trainset to do the classification with our FCN. Using the Product Images induces lot of false positives. The Product Images have two parts - the brand logo and the warning labels. The former is the only discriminative part as the latter is common to all brands. <<Since our dataset contains only 5 unique products, that information isn't reflected in the trainset>>. The network is made to think that certain warning labels are specific to each product. This is the reason why using Product Images results in lot of false positives. Hence, we chose to use the Brand Images instead.

For background patches, we took a very small amount of (6.49\%) images from the Shelf Images, containing all 11 images from one shelf and all 12 images from a different shelf. We make sure that there's no overlap in the shelves in our train and test set. The exact details are given in Appendix. From these images, we extract background patches which used as the negative class in our classification task. We randomly use 20\% of this trainset to be our validation set.

A synthetic dataset is created to train RefineNet ConvAE by using product images in train set and following procedure detailed earlier. 

\subsection{Test set}
The testing task is the object detection task. The Shelf Images excluding the ones used for background patches, are used as the test set. This comes around to 331 test images. The annotations are provided for all instances of the products in each shelf image.

\subsection{Previous Work}

An initial work on this dataset is done by \cite{cigarette}. They first try to segment all the products by calculating number of shelves and applying a height and width constraint from the dataset. This works well in segmenting all the boxes as the shape of products in this dataset is rather constant. They also classify each product image into given classes. But they do not combine both of them to give a working object detection method.

We don't have any other previous work to compare with, so we make a simple baseline which is described as follows. We employ a sliding window of different scales, aspect ratios on the test image and classify the sliding window into different classes. We then use non max suppression (NMS) to remove the redundant bounding boxes. The classifier used here is the same FCN which is used in other results as well. This baseline would show the significance of the image pyramid testing methodology as well as the significance of ConvAE.

\subsection{Implementation Details}

For the \textbf{FCN}, we set the last convolutional layer kernel size as (2,4). We train the FCN using SGD with momentum 0.9 and learning rate 1e-3. We use weight decay of 5e-4 and an early stopping of 30 epochs. For creating the \textbf{Synthetic Dataset}, we set the image height and width as (2000,3000) and (1200,2000) according to number of shelves (rows) in the image. For the \textbf{ConvAE}, we set the number of filters in the 3 conv and deconv blocks as 16,24,32 respectively. We train the ConvAE using SGD with momentum 0.9 and learning rate 1e-3. We use weight decay of 5e-4 and early stopping of 5 epochs. 

\subsection{Results}
\label{cig:results}

The mAP of different models are shown in Table 1. The mAP is calculated as the mean of Average Precision (AP) of all classes. The calculation of AP is taken directly from VOC devkit (Matlab\footnote{\url{http://host.robots.ox.ac.uk/pascal/VOC/voc2012/\#devkit}}, Python\footnote{\url{https://github.com/rbgirshick/py-faster-rcnn/blob/master/lib/datasets/voc\_eval.py}}). We use the VOC07 11 point metric.

\begin{table}[]
\centering
\caption{Object Detection results on Cigarette Dataset}
\label{table:cigmap}
\begin{tabular}{|l|l|}
\hline
Model                                  & mAP (iou=0.1) \\ \hline
Selective Search (with FCN Classifier) & -             \\ \hline
Sliding Window (with FCN Classifier)   & 0.278         \\ \hline
Pyramid FCN Model without AE           & 0.2805        \\ \hline
Pyramid FCN Model with ConvAE         & 0.3069        \\ \hline
\end{tabular}
\end{table}

\begin{figure*}
\includegraphics[width=0.49\textwidth, height=0.3\textheight]{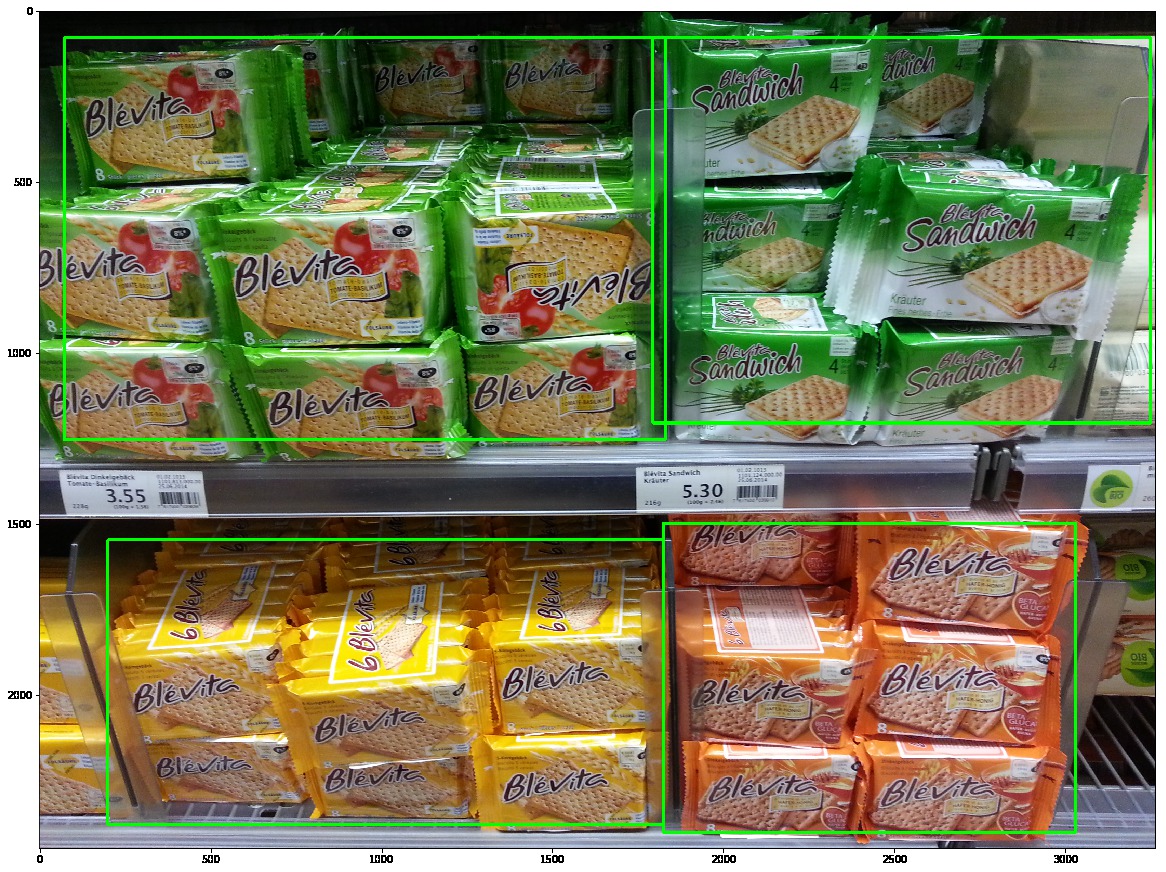}
\includegraphics[width=0.49\textwidth, height=0.3\textheight]{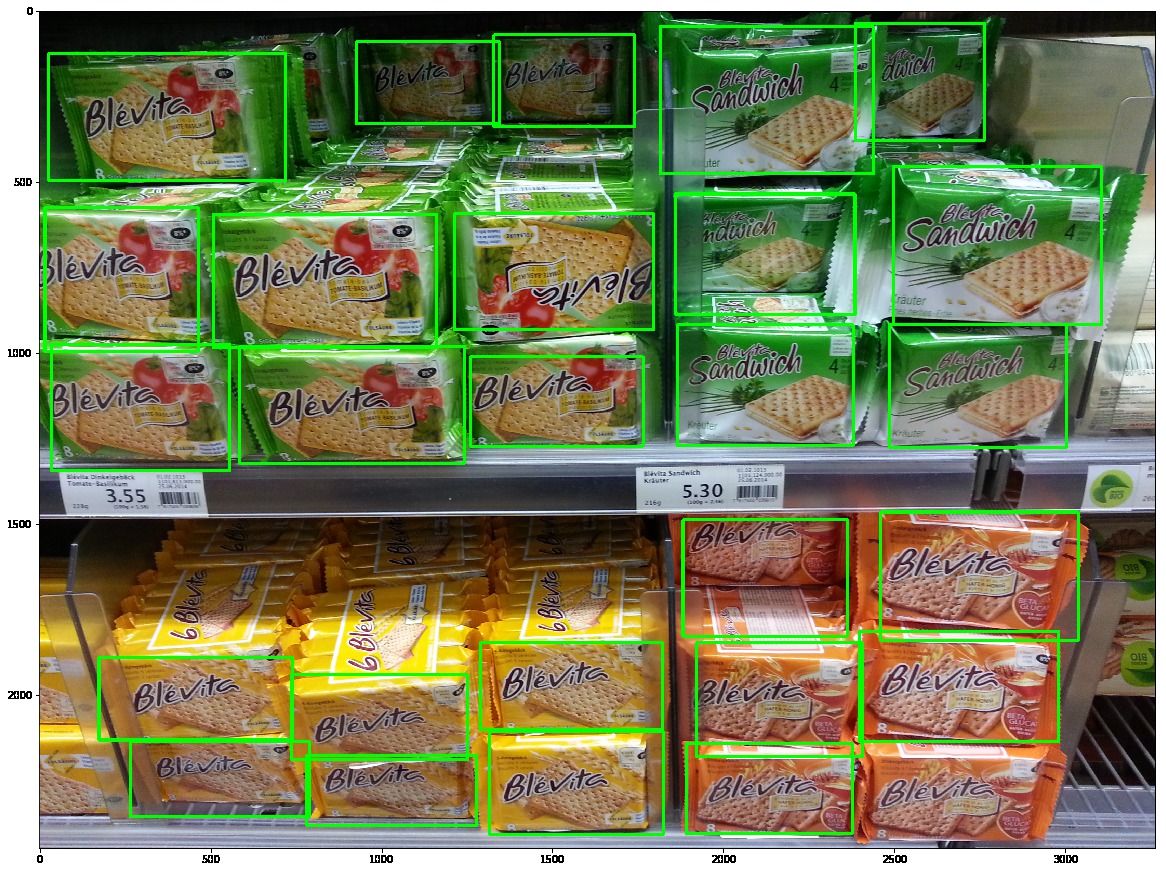}
\caption{Annotations from \cite{grocerydset} (left) and \cite{planogram} (right)}
\label{fig:grocery}
\end{figure*}

For the baseline method, we used 3 scales and 5 aspect ratios for the sliding windows. The feature for each patch was extracted using the PyramidFCN. Our image pyramid method performs better than the baseline sliding window method. The results get better when we add the refinement module to the FCN model (Table \ref{table:cigmap}). The difference in accuracy is less due to the fixed structure of objects in the cigarette dataset. This gives the sliding window method an extra knowledge and allows it to get a decent accuracy. We show later in Section \ref{grocery:results} that this method fails badly in case of a varied dataset which has different object shapes and sizes, while our image pyramid method works much better\footnote{Note that these are absolute values, not percentages.}.

We also tried to compare our method with another baseline by taking region proposals from selective search \cite{ssearch} and classifying it with our classifier. But performing the selective search on each image took around 10 minutes and would've taken almost 3 days to test on the entire testset. Changing parameters to lower the time would result in unfavourable region proposals. Hence we tried it on a subset and saw very poor performance, in the order of 1e-4 mAP. Hence, we decided not to test on the entire dataset. The aim was to compare against the classic region proposal methods \cite{rcnn}. Note that we can't use fully supervised methods like RPN \cite{rpn}.

\section{Grocery Dataset}

\subsection{Dataset Description}

A supermarket dataset containing 8350 training images spanning 80 semantic object categories was introduced by \cite{grocerydset}. The training images are taken under ideal, studio conditions. Their test set contains 680 real shelf images taken in different grocery stores. 

The annotation on the test set covers a group of individual products \textbf{as opposed to} giving bounding boxes for each instance following the object detection task. Fortunately, \cite{planogram} released the bounding box annotation for a subset (74 images) of the test set. We use this as our test set. The comparison between the annotations is shown in \ref{fig:grocery}. This subset contains 12 classes. Hence, we train our FCN and the ConvAE only on these 12 classes. We observe a similar trend on the full testset but are unable to measure it due to lack of annotations.

\subsection{Previous Work}

A method was proposed by \cite{grocerydset} to recognize products on this dataset, but it doesn't perform the object detection task that we aim to do. Their method consists of 3 steps: Multi-class ranking, Fast Dense Pixel Matching and a genetic algorithm based multi-label image classification. The evaluation results shown based on the group level annotations shown in Figure \ref{fig:grocery} (Left). \cite{planogram} have also worked on the same dataset, but their focus was on planogram compliance. They're given the structure of a planogram (called reference planogram) along with the objects it contains and are asked to check whether the observed planogram is compliant to the planned (reference) one. They also detect missing or misplaced items. We are unable to find any suitable previous benchmark, so we follow the same baseline as described in Section \ref{cig:results}

\subsection{Implementation Details}

For the FCN, we set the last convolutional layer kernel size as (7,5). The input image is of fixed dimensions taken from average of the dataset. We train the FCN using SGD with momentum 0.9 and learning rate 1e-3. We use weight decay of 5e-4 and an early stopping of 30 epochs. We used common data augmentation techniques like flipping, scaling, rotations using imgaug package\footnote{\url{https://github.com/aleju/imgaug}}

We tried using the Refine-net here, but it wasn't helpful as the FCN isn't very strong in generating the feature maps without false-positives. This is mentioned in Section \ref{observation}. So, we report the results directly on outputs of FCN, which is thresholded and converted to bounding boxes (Figure \ref{fig:all_ops} without the refinement step).

\subsection{Results}
\label{grocery:results}

We follow the same evaluation metric as described above in Section \ref{cig:results}. As mentioned in Section \ref{cig:results}, one can see that the sliding window method with multiple scales, aspect ratios and NMS fails to work in the case where the objects don't have a fixed aspect ratio or size. For selective search model, we face a similar problem as well as similar performance (explained in Section \ref{cig:results}). The image pyramid testing method using our FCN works much better than the baseline (Table \ref{table:grocerymap}). The results show that the method works well even for generalized datasets which has semantic grocery categories containing only one or few instances of each product.

\begin{table}[]
\centering
\caption{Object Detection results on Grocery Dataset}
\label{table:grocerymap}
\begin{tabular}{|l|l|}
\hline
Model                                  & mAP (iou=0.1) \\ \hline
Selective Search (with FCN Classifier) & -             \\ \hline
Sliding Window (with FCN Classifier)   & 0.0249        \\ \hline
Pyramid FCN Model without AE           & 0.2801        \\ \hline
\end{tabular}
\end{table}

\begin{figure*}
\includegraphics[width=0.49\textwidth, height=0.2\textheight]{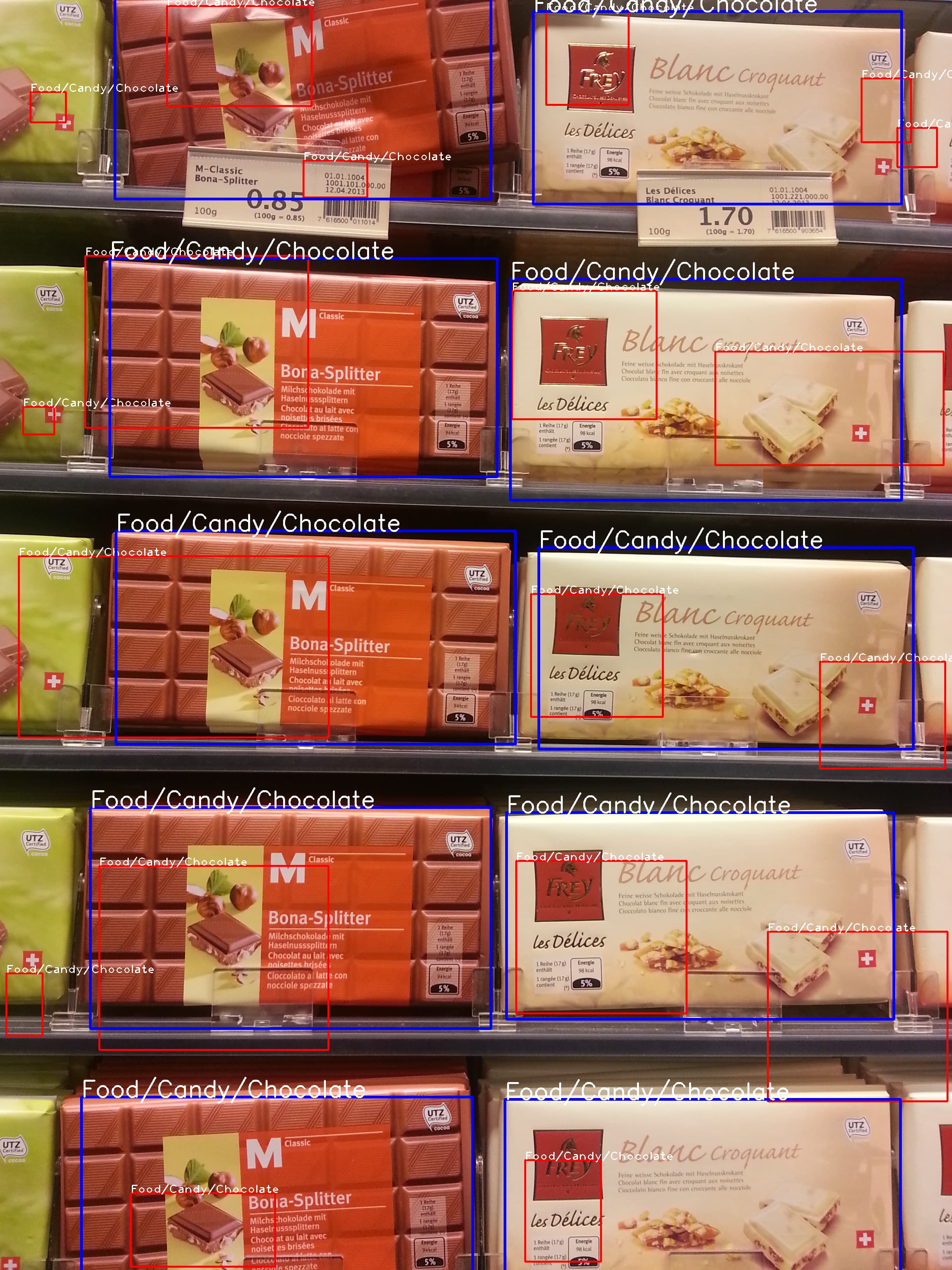}
\includegraphics[width=0.49\textwidth, height=0.2\textheight]{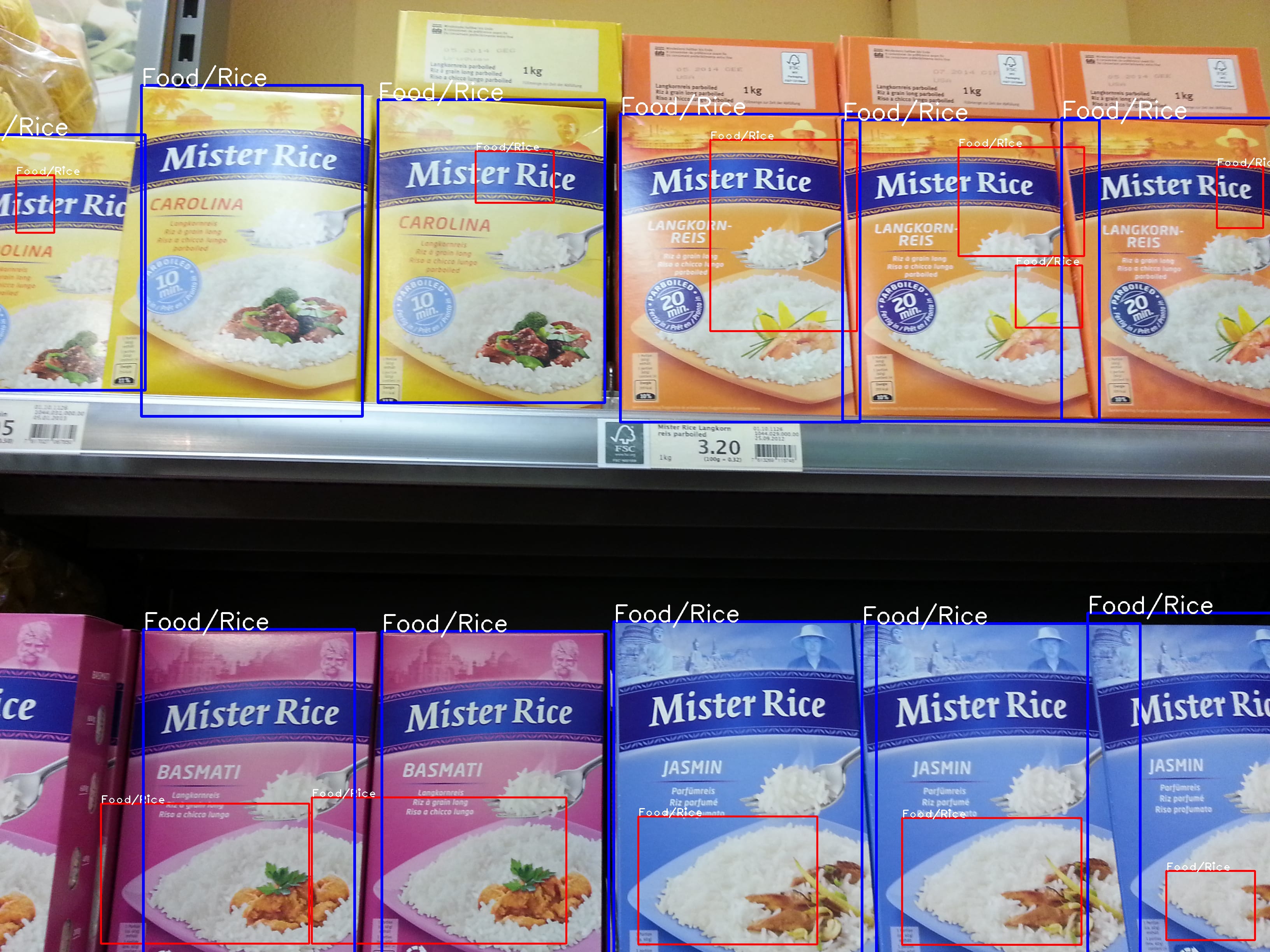}
\caption{Sample Results of Pyramid FCN Model without AE. Blue shows ground-truth boxes while Red shows model's prediction}
\label{fig:grocresult}
\end{figure*}

\subsection{GroZi-120 dataset}

Grozi-120 dataset \cite{grozi120} contains 120 specific product instances as training set. The test-set contains 29 videos of shelves containing these 120 products. \cite{grocerydset} had selected 885 frames from these 29 videos and released the annotations to show their results. Unfortunately these annotations aren't bounding boxes but just list of classes present in the frame. It doesn't give the bounding boxes of all the objects present in the frame. Hence, we were unable to show the results on this dataset.  

\subsection{Discussion}

We note that the IoU threshold for calculating the mAP is low (0.1), this is due to two factors. First, the weakly supervised method is unable to fit the entire object. Second, it gives a combined bounding box in the cases where the boundary between similar objects is very hard to distinguish. The latter is unique to shelf datasets and shouldn't be a limitation of the method in general datasets. The former can be tackled as well. Our refinement module is complementary to other methods like \cite{hidenseek} which helps to improve the extent of the localization and we expect our method's performance to increase when combined with such methods. Future work can be focused on improving the method's performance on higher thresholds. Another thing to note is that, our method is more suited to structured object instances which is not necessarily the case with other WSOL methods. To make our method more robust to any object, one should collect training instances of the object spanning different views, orientations etc.

\section{Conclusion}

We have proposed a novel and extremely simple method to do object localization using just the object instances (classification labels). We have also proposed an optional refinement module to increase the extent of localization and to reduce false positives as well. We have implemented it on relevant datasets and shown good performance on all of them. We've also established the superior performance of the method on a varied and general dataset as well.

\bibliographystyle{ACM-Reference-Format}
\bibliography{egbib}

\end{document}